\documentclass[10pt,letterpaper,oneside,twocolumn]{IEEEtran}

\makeatletter
\def\@endtheorem{\endtrivlist}
\makeatother

\usepackage{graphicx}
\usepackage{amsmath}
\usepackage{amsfonts}
\usepackage{epsfig}
\usepackage{url}
\usepackage{cite}
\usepackage{algorithm}
\usepackage{algorithmic}

\newcommand{\BibTeX}{{\rmfamily B\kern-.05em{\scshape i\kern-.025em
b}\kern-.08em T\kern-.1667em\lower.7ex\hbox{E}\kern-.125emX}}

\begin{document}

\title{A Leaf Recognition Algorithm for Plant Classification Using Probabilistic Neural Network}

\author{
Stephen Gang Wu$^1$,  Forrest Sheng Bao$^2$, Eric You Xu$^3$, Yu-Xuan Wang$^4$, Yi-Fan Chang$^5$ and Qiao-Liang Xiang$^4$\\
$^1$ Institute of Applied Chemistry, Chinese Academy of Science, P. R. China \\
$^2$ Dept. of Computer Science, Texas Tech University, USA \\
$^3$ Dept. of Computer Science \& Engineering,  Washington University in St. Louis, USA \\ %
$^4$ School of Information \& Telecommunications Eng., Nanjing Univ. of P \& T,  P. R. China \\
$^5$ Dept. of Electronic Engineering,  National Taiwan Univ. of Science \& Technology, Taiwan, R. O. China \\

Corresponding author's E-mail : \url{shengbao@ieee.org}}
\maketitle

\begin{abstract}
In this paper, we employ Probabilistic Neural Network (PNN) with image and data processing techniques to implement a general purpose automated leaf recognition for plant classification. 
12 leaf features are extracted and orthogonalized into 5 principal variables which consist the input vector of the PNN. The PNN is trained by 1800 leaves to classify 32 kinds of plants with an accuracy greater than 90\%. Compared with other approaches, our algorithm is an accurate artificial intelligence approach which is fast in execution and easy in implementation.
\end{abstract}



\section{Introduction}
Plants exist everywhere we live, as well as places without us. Many of them carry significant information for the development of human society. The urgent situation is that many plants are at the risk of extinction. So it is very necessary to set up a database 
for plant protection
 \cite{Du2007, A_Compuerized_Plant_Species_Recognition_System, An_OOPR-based_rose_variety_recognition_system, Leaf_shape_analysis_using_the_multiscale_Minkowski_fractal_dimension}. We believe that the first step is to teach a computer how to classify plants.

Compared with other methods, such as cell and molecule biology methods, classification based on leaf image is the first choice for leaf plant classification. Sampling leaves and photoing them are low-cost and convenient. One can easily transfer the leaf image to a computer and a computer can extract features automatically in image processing techniques. 

Some systems employ descriptions used by botanists\cite{DELTA-overview, Machine_learning_techniques_for_ontology_based_leaf_classification, Automated_Leaf_Shape_Description_for_Variety_Testing_in_Chrysanthemums, SPIE2345}.
But it is not easy to extract and transfer those features to a computer automatically. This paper tries to prevent human interference in feature extraction. 

It is also a long discussed topic on how to extract or measure leaf features \cite{A_Leaf_Vein_Extraction_Method_Based_On_Snakes_Technique, Combined_thresholding_and_neural_network_approach_for_vein_pattern_extraction_from_leaf_images, ELIS, A_TWO-STAGE_APPROACH_FOR_LEAF_VEIN_EXTRACTION, Shape_based_leaf_image_retrieval, ICMLC2003,  Interactive_venation-based_leaf_shape_modeling}. That makes the application of pattern recognition in this field a new challenge\cite{Du2007}\cite{Biometry}. According to \cite{Du2007}, data acquisition from living plant automatically by the computer has not been implemented.




Several other approaches used their pre-defined features. \textit{Miao et al.} proposed an evidence-theory-based rose classification \cite{An_OOPR-based_rose_variety_recognition_system} based on many features of roses. \textit{Gu et al.} tried leaf recognition using skeleton segmentation by wavelet transform and Gaussian interpolation \cite{Leaf_Recognition_Based_on_the_Combination_of_Wavelet_Transform_and_Gaussian_Interpolation}. 
 \textit{Wang et al.} used a moving median center (MMC) hypersphere classifier \cite{Recognition_of_Leaf_Images_Based_on_Shape_Features_Using_a_Hypersphere_Classifier}. Similar method was proposed by \textit{Du et al.}\cite{Du2007}. Their another paper proposed a modified dynamic programming algorithm for leaf shape matching\cite{DuIMC}. \textit{Ye et al.} compared the similarity between features to classify plants \cite{A_Compuerized_Plant_Species_Recognition_System}. 

Many approaches above employ $k$-nearest neighbor ($k$-NN) classifier\cite{Du2007}\cite{Leaf_Recognition_Based_on_the_Combination_of_Wavelet_Transform_and_Gaussian_Interpolation}\cite{ Recognition_of_Leaf_Images_Based_on_Shape_Features_Using_a_Hypersphere_Classifier} while some papers adopted Artificial Neural Network (ANN). \textit{Saitoh et al.} combined flower and leaf information to classify wild flowers\cite{wild_flower}. \textit{Heymans et al.} proposed an application of ANN to classify opuntia species \cite{A_Neural_Network_for_Opuntia_leaf-form_recognition}.  \textit{Du et al.} introduced shape recognition based on radial basis probabilistic neural network
which is trained by orthogonal least square algorithm (OLSA) and optimized by recursive OLSA 
\cite{Du_ISNN}. It performs plant
recognition through modified Fourier descriptors of leaf shape. 






Previous work have some disadvantages. Some are only applicable to certain species\cite{An_OOPR-based_rose_variety_recognition_system} \cite{Biometry} \cite{A_Neural_Network_for_Opuntia_leaf-form_recognition}. As expert system, some methods compare the similarity between features\cite{A_Compuerized_Plant_Species_Recognition_System}\cite{SPIE2345}. It requires pre-process work of human to enter keys manually. This problem also happens on methods extracting features used by botanists\cite{Automated_Leaf_Shape_Description_for_Variety_Testing_in_Chrysanthemums}\cite{Biometry}. 

Among all approaches, ANN has the fastest speed and best accuracy for classification work. \cite{Du_ISNN} indicates that ANN classifiers (MLPN, BPNN, RBFNN and RBPNN) run faster than k-NN (k=1, 4) and MMC hypersphere classifier while ANN classifiers advance other classifiers on accuracy. So this paper adopts an ANN approach.

This paper implements a leaf recognition algorithm using easy-to-extract features and high efficient recognition algorithm. Our main improvements are on feature extraction and the classifier. All features are extracted from digital leaf image. Except one feature, all features can be extracted automatically. 12 features are orthogonalized by Principal Components Analysis (PCA)\cite{PCA-UCSD}. As to the classifier, we use PNN\cite{PNN} for its fast speed and simple structure. 
The whole algorithm is easy-to-implement, using common approaches.

\setlength{\unitlength}{0.5cm}
\begin{figure}[!hbt]
\begin{center}
\label{diagram}
 \begin{picture}(6,13)
\put(-1,0){\framebox(8,1){Display \& Compare results}}
\put(3,2){\vector(0,-1){1}}
\put(0,2){\framebox(6,1){Test PNN}}
\put(3,4){\vector(0,-1){1}}
\put(0,4){\framebox(6,1){Train PNN}}
\put(3,6){\vector(0,-1){1}}
\put(0,6){\framebox(6,1){PCA}}
\put(3,8){\vector(0,-1){1}}
\put(0,8){\framebox(6,1){extract features}}
\put(3,10){\vector(0,-1){1}}
\put(0,10){\framebox(6,1){Process Image}}
\put(3,12){\vector(0,-1){1}}
\put(-1,12){\framebox(8,1){Capture digital leaf image}}
\end{picture}
\caption[fig]{Flow diagram of proposed scheme}
\end{center}
\end{figure}
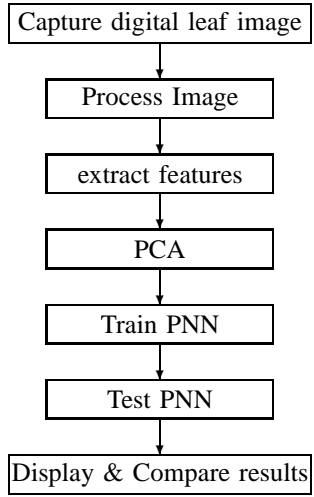

The rest of this paper is organized as follows. Sec. \ref{image_processing} discusses image pre-processing. Sec. \ref{feature_extraction} introduces how 12 leaf features are extracted. PCA and PNN are discussed in Sec. \ref{scheme}. Experimental results are given in Sec. \ref{result}. Future work on improving our algorithm is mentioned in Sec. \ref{futurework}. Sec. \ref{conclusion} concludes this paper. 

\section{Image Pre-processing}
\label{image_processing}
\subsection{Converting RGB image to binary image}

The leaf image is acquired by scanners or digital cameras. Since we have not found any digitizing device to save the image in a lossless compression format, the image format here is JPEG. 
All leaf images are in 800 x 600 resolution. There is no restriction on the direction of leaves when photoing.

An RGB image is firstly converted into a grayscale image. Eq. \ref{rgb2gray} is the formula used to convert RGB value of a pixel into its grayscale value. 
\begin{equation}
 gray = 0.2989 * R + 0.5870 * G + 0.1140 * B 
\label{rgb2gray}
\end{equation}
where R, G, B correspond to the color of the pixel, respectively.

\begin{figure}[!hbt]
\includegraphics[scale=0.41]{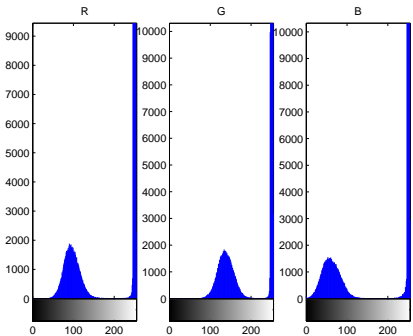}
\centering \caption[fig]{RGB histogram}
\label{RGB}
\end{figure}

The level to convert grayscale into binary image is determined according to the RGB histogram. We accumulate the pixel values to color R, G, B respectively for 3000 leaves and divide them  by 3000, the number of leaves. The average histogram to RGB of 3000 leaf images is shown as Fig. \ref{RGB}.

There are two peaks in every color's histogram. The left peak refers to pixels consisting the leaf while the right peak refers to pixels consisting the white background. The lowest point between two peaks is around the value 242 on the average. So we choose the level as 0.95 (242/255=0.949). The output image replaces all pixels in the input image with luminance greater than the level by the value 1 and replaces all other pixels by the value 0. 

A rectangular averaging filter of size  $3 \times 3$ is applied to filter noises. Then pixel values are rounded to 0 or 1.

\subsection{Boundary Enhancement}
When mentioning the leaf shape, the first thing appears in your mind might be the margin of a leaf. Convolving the image with a Laplacian filter of following $3 \times 3$ spatial mask:
\[
\begin{matrix}
 0 & 1 & 0 \\
1 & -4 & 1\\
0 & 1 & 0
\end{matrix}
\]
we can have the margin of the leaf image.

\begin{figure}[!hbt]
\begin{center}
\includegraphics[scale=0.5]{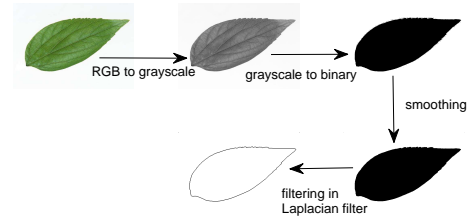}
\caption[fig]{A pre-processing example}
\label{binary}
\end{center}
\end{figure}

 An example of image pre-processing is illustrated in Fig. \ref{binary}. To make boundary as a black curve on white background, the ``0'' ``1'' value of pixels is swapped. 

\section{Feature Extraction}
\label{feature_extraction}
In this paper, 12 commonly used digital morphological features (DMFs), derived from 5 basic features, are extracted so that a computer can obtain feature values quickly and automatically (only one exception).

\subsection{Basic Geometric Features}
Firstly, we obtain 5 basic geometric features. 

\subsubsection{Diameter}
The diameter is defined as the longest distance between any two points on the margin of the leaf.
It is denoted as $D$.

\subsubsection{Physiological Length}
The only human interfered part of our algorithm is that you need to mark the two terminals of the main vein of the leaf via mouse click. 
The distance between the two terminals is defined as the physiological length.
It is denoted as $L_{p}$.

\subsubsection{Physiological Width}
Drawing a line passing through the two terminals of the main vein, one can plot infinite lines orthogonal to that line. The number of intersection pairs between those lines and the leaf margin is also infinite. The longest distance between points of those intersection pairs is defined at the physiological width. It is denoted as $W_{p}$.

Since the coordinates of pixels are discrete, we consider two lines are orthogonal if 
their degree is $90^{\circ}\pm 0.5^{\circ}$.

\begin{figure}[!hbt]
\begin{center}
\includegraphics[scale=0.4]{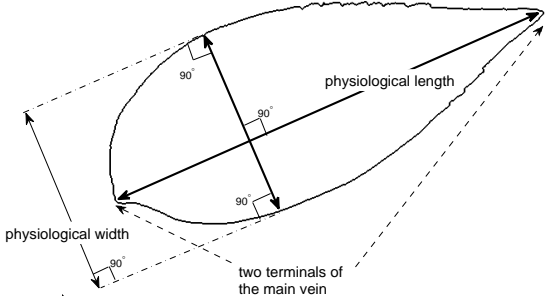}
\caption[fig]{Relationship between Physiological Length and Physiological Width}
\label{longest}
\end{center}
\end{figure}

The relationship between physiological length and physiological width is illustrated in Fig. \ref{longest}.

\subsubsection{Leaf Area}
The value of leaf area is easy to evaluate, just counting the number of pixels of binary value 1 on smoothed leaf image. 
It is denoted as $A$.

\subsubsection{Leaf Perimeter}
Denoted as $P$, leaf perimeter is calculated by counting the number of pixels consisting leaf margin.

\subsection{12 Digital Morphological Features}
Based on 5 basic features introduced previously, we can define 12 digital morphological features used for leaf recognition.

\subsubsection{Smooth factor}
We use the effect of noises to image area to describe the smoothness of leaf image. In this paper, smooth factor is defined as the ratio between area of leaf image smoothed by $5 \times 5$ rectangular averaging filter and the one smoothed by $2 \times 2$ rectangular averaging filter.

\subsubsection{Aspect ratio}
The aspect ratio is defined as the ratio of physiological length $L_{p}$ to physiological width $W_{p}$, thus $L_{p}/W_{p}$.

\subsubsection{Form factor}
This feature is used to describe the difference between a leaf and a circle.
It is defined as 
${4\pi A}/{P^2}$,
where $A$ is the leaf area and $P$ is the perimeter of the leaf margin. 

\subsubsection{Rectangularity}
Rectangularity describes the similarity between a leaf and a rectangle. It is defined as $L_{p}W_{p}/{A}$,
where $L_{p}$ is the physiological length, $W_{p}$ is the physiological width and $A$ is the leaf area.

\subsubsection{Narrow factor}
Narrow factor is defined as the ratio of the diameter $D$ and physiological length $L_{p}$, thus $D/L_{p}$.

\subsubsection{Perimeter ratio of diameter}
Ratio of perimeter to diameter, representing the ratio of leaf perimeter $P$ and leaf diameter $D$, is calculated by $P/D$.

\subsubsection{Perimeter ratio of physiological length and physiological width}
This feature is defined as the ratio of leaf perimeter $P$ and the sum of physiological length $L_{p}$ and physiological width $W_{p}$, thus $P/(L_{p}+W_{p})$.

\subsubsection{Vein features}
We perform morphological opening\cite{Gonzalez} on grayscale image with falt, disk-shaped structuring element of radius 1,2,3,4 and substract remained image by the margin. The results look like the vein. That is why following 5 feature are called vein features. Areas of left pixels are denoted as $A_{v1}$, $A_{v2}$, $A_{v3}$ and $A_{v4}$ respectively. Then we obtain the last 5 features: $A_{v1}/A$, $A_{v2}/A$, $A_{v3}/A$, $A_{v4}/A$, $A_{v4}/A_{v1}$.

%
%
%

Now we have finished the step of feature acquisition and go on to the data analysis section.

\section{Proposed Scheme}
\label{scheme}

\subsection{Principal Component Analysis (PCA)}
\label{dataanalysis}

%


To reduce the dimension of input vector of neural network, PCA is used to orthogonalize 12 features. The purpose of PCA is to present the information of original data as the linear combination of certain linear irrelevant variables.
Mathematically, PCA transforms the data to a new coordinate system such that the greatest variance by any projection of the data comes to lie on the first coordinate, the second greatest variance on the second coordinate, and so on. Each coordinate is called a principal component. 

In this paper, the contribution of first 5 principal components is 93.6\%.
To balance the computational complexity and accuracy, we adopt 5 principal components.

 When using our algorithm, one can use the mapping $f: \mathbb{R}^{12} \rightarrow \mathbb{R}^{5}$ to obtain the values of components in the new coordinate system.

\subsection{Introduction to Probabilistic Neural Network}
\label{PNN}
An artificial neural network (ANN) is an interconnected group of artificial neurons simulating the thinking process of human brain.
One can consider an ANN as a ``magical'' black box trained to achieve expected intelligent process, against the input and output information stream. Thus, there is no need for a specified algorithm on how to identify different plants. 
PNN is derived from Radial Basis Function (RBF) Network which is an ANN using RBF. RBF is a bell shape function that scales the variable nonlinearly.

PNN is adopted for it has many advantages \cite{PNN-Good}. Its training speed is many times faster than a BP network. PNN can approach a Bayes optimal result under certain easily met conditions\cite{PNN}. Additionally, it is robust to noise examples. We choose it also for its simple structure and training manner.

The most important advantage of PNN is that training is easy and instantaneous\cite{PNN}. Weights are not ``trained'' but assigned. Existing weights will never be alternated but only new vectors are inserted into weight matrices when training. 
So it can be used in real-time.
Since the training and running procedure can be implemented by matrix manipulation, the speed of PNN is very fast.
 
The network classfies input vector into a specific class because that class has the maximum probability to be correct. In this paper, the PNN has three layers: the Input layer, Radial Basis Layer and the Competitive  Layer. 
Radial Basis Layer evaluates vector distances between input vector and row weight vectors in weight matrix. These distances are scaled by Radial Basis Function nonlinearly. Then the Competitive Layer finds the shortest distance among them, and thus finds the training pattern closest to the input pattern based on their distance.


\setlength{\unitlength}{0.5cm}
 \begin{figure*}[!htb]
 \begin{center}
 \begin{picture}(25,9.1)
\put(1,0.5){\oval(2,1)[b]}
\put(1,8.1){\oval(2,1)[t]}
\put(-0.5,9){Input Layer}
 \put(1,1){\makebox(0,0){\small{\textit{R}}}}
 \put(1,2.5){\makebox(0,2){\rule{3mm}{20mm}}}
 \put(1,5){\vector(1,0){3}}
 \put(3,5){\makebox(0,0.6){\textbf{P}}}
 \put(3,5){\makebox(0,-0.7){\small{\textit{R}$\times $\small{\textit{1}}}}}
\put(9.5,0.5){\oval(14,1)[b]}
\put(9.5,8.1){\oval(14,1)[t]}
\put(7,9){Radial Basis Layer}
 \put(3,2){\makebox(0,0){1}}
 \put(3.5,2){\vector(1,0){2.5}}
 \put(6,1.5){\framebox(2,1){\textbf{b}}}
 \put(5,7){\framebox(2,1){$ \mathbf{W} $}}
 \put(5,6.5){\makebox(0,0){\small{\textit{Q}}$\times $\small{\textit{R}}}}
\put(4,4.5){\framebox(4,1){$\| \mathbf{W} - \mathbf{p}\|$}}
\put(6,7){\vector(0,-1){1.5}}
\put(6,1){\makebox(0,0){\small{\textit{Q}}$\times $\small{\textit{1}}}}
\put(8,2){\line(1,0){1}}
\put(8,5){\line(1,0){1}}
\put(11,3.5){\circle{1}}
\put(10.8,3.5){\makebox(0,0){\large{ $\cdot \ast$}}}
\put(9,5){\vector(2,-1){2}}
\put(9,2){\vector(2,1){2}}
\put(11.5,3.5){\vector(1,0){2}}
\put(12.5,4){\makebox(0,0){$\mathbf{n}$}}
\put(12.4,3){\makebox(0,0){\small{\textit{Q}}$\times$ 1}}
\put(13.5,1.5){\framebox(2,4){\epsfig{file=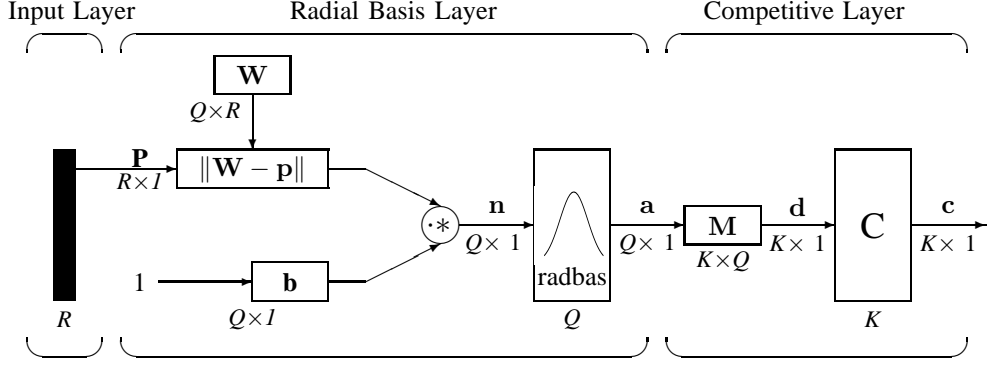,width=29pt,height=30pt}}}
\put(14.5,2.4){\makebox(0,-0.5){radbas}}
\put(14.5,1){\makebox(0,0){\small{\textit{Q}}}}
\put(15.5,3.5){\vector(1,0){2}}
\put(16.5,4){\makebox(0,0){$\mathbf{a}$}}
\put(16.5,3){\makebox(0,0){\small{\textit{Q}}$\times$ 1}}
\put(17.5,3){\framebox(2,1){$ \mathbf{M} $}}
\put(17.8,2.4){\small{\textit{K}}$\times$\small{\textit{Q}}}
\put(19.5,3.5){\vector(1,0){2}}
\put(20.5,4){\makebox(0,0){$\mathbf{d}$}}
\put(20.5,3){\makebox(0,0){\small{\textit{K}}$\times$ 1}}
\put(21.5,1.5){\framebox(2,4){\Large{C}}}
\put(22.5,1){\makebox(0,0){\small{\textit{K}}}}
\put(23.5,3.5){\vector(1,0){2}}
\put(24.5,4){\makebox(0,0){$\mathbf{c}$}}
\put(24.5,3){\makebox(0,0){\small{\textit{K}}$\times$ 1}}
\put(21,0.5){\oval(8,1)[b]}
\put(21,8.1){\oval(8,1)[t]}
\put(18,9){Competitive Layer}
\end{picture}
\centering \caption[fig]{Network Structure, R=5, Q=1800, K=32}
\label{structure}
\end{center}
\end{figure*}

\subsection{Network Structure}
The network structure in our purposed scheme is illustrated in Fig. \ref{structure}. We adopt symbols and notations used in the book \textit{Neural Network Design}\cite{NNdesign}. These symbols and notations are also used by MATLAB Neural Network Toolbox\cite{NNToolbox}. Dimensions of arrays are marked under their names.
\subsubsection{Input Layer}
The input vector, denoted as $\mathbf{p}$, is presented as the black vertical bar in Fig. \ref{structure}. Its dimension is $R \times 1$. In this paper, $R=5$.

\subsubsection{Radial Basis Layer}
In Radial Basis Layer, the vector distances between input vector $\mathbf{p}$ and the weight vector made of each row of weight matrix $\mathbf{W}$ are calculated. 
 Here, the vector distance is defined as the dot product between two vectors\cite{Specht1988}. Assume the dimension of  $\mathbf{W}$  is $Q \times R$. The dot product between $\mathbf{p}$ and the $i$-th row of $\mathbf{W}$ produces the $i$-th element of the distance vector $||\mathbf{W}  - \mathbf{p} ||$, whose dimension is $Q \times 1$, as shown in Fig. \ref{structure}. The minus symbol, ``$-$", indicates that it is the distance between vectors.



Then, the bias vector $\mathbf{b}$ is combined with $||\mathbf{W} - \mathbf{p}||$ by an element-by-element multiplication, represented as ``$\cdot \ast$" in Fig. \ref{structure}. The result is denoted as $\mathbf{n} = ||\mathbf{W} - \mathbf{p}|| \cdot \ast \mathbf{p}$.



The transfer function in PNN has built into a distance criterion with respect to a center. In this paper, we define it as
\begin{equation}
 radbas(n) = e ^{-{n^2}}
\label{radbas}
\end{equation}
Each element of $\mathbf{n}$ is substituted into Eq. \ref{radbas} and produces corresponding element of $\mathbf{a}$, the output vector of Radial Basis Layer. We can represent the \textit{i}-th element of $\mathbf{a}$ as 
\begin{equation}
 a_{i}=radbas(||\mathbf{W}_{i} - \mathbf{p}||\cdot \ast \mathbf{b}_{i})
\end{equation}
where $\mathbf{W}_{i}$ is the vector made of the $i$-th row of $\mathbf{W}$ and $\mathbf{b}_{i}$ is the $i$-th element of bias vector $\mathbf{b}$.

\subsubsection{Some characteristics of Radial Basis Layer}
The $i$-th element of $\mathbf{a}$ equals to 1 if the input $\mathbf{p}$ is identical to the $i$-th row of input weight matrix $\mathbf{W}$. A radial basis neuron with a weight vector close to the input vector $\mathbf{p}$ produces a value near 1 and then its output weights in the competitive layer will pass their values to the competitive function which will be discussed later. It is also possible that several elements of $\mathbf{a}$ are close to 1 since the input pattern is close to several training patterns.


\subsubsection{Competitive Layer}
There is no bias in Competitive Layer. In Competitive Layer, the vector $\mathbf{a}$ is firstly multiplied with layer weight matrix $\mathbf{M}$, producing an output vector $\mathbf{d}$. The competitive function, denoted as $\mathbf{C}$ in Fig. \ref{structure}, produces a 1 corresponding to the largest element of $\mathbf{d}$, and 0's elsewhere. The output vector of competitive function is denoted as $\mathbf{c}$. The index of 1 in $\mathbf{c}$ is the number of plant that our system can classify. It can be used as the index to look for the scientific name of this plant. The dimension of output vector, $K$, is 32 in this paper.

\subsection{Network Training}
Totally 1800 pure leaves are sampled to train this network. Those leaves are sampled in the campus of Nanjing University and Sun Yat-Sen arboretum, Nanking, China. Most of them are common plants in Yangtze Delta, China. Details about the leaf numbers of different kinds of plants are given in Table \ref{training}. The reason why we sample different pieces of leaves to different plants is the difficulty to sample leaves varies on plants.

\subsubsection{Radial Basis Layer Weights} $\mathbf{W}$ is set to the transpose of $R \times Q$ matrix of $Q$ training vectors. Each row of $\mathbf{W}$ consists of 5 principal variables of one trainging samples. Since 1800 samples are used for training, $Q=1800$ in this paper.

\subsubsection{Radial Basis Layer Biases} All biases in radial basis layer are all set to $\sqrt{\ln{0.5}}/s$ resulting in radial basis functions that cross 0.5 at weighted inputs of $\pm s$.  $s$ is called the spread constant of PNN.

The value of $s$ can not be selected arbitrarily. Each neuron in radial basis layer will respond with 0.5 or more to any input vectors within a vector distance of $s$ from their weight vector. A too small $s$ value can result in a solution that does not generalize from the input/target vectors used in the design. In contrast, if the spread constant is large enough, the radial basis neurons will output large values (near 1.0) for all the inputs used to design the network.

In this paper, the $s$ is set to 0.03($\simeq 1/32$) according to our experience.

\subsubsection{Competitive Layer Weights}
 $\mathbf{M}$ is set to $K \times Q$ matrix of $Q$ target class vectors. The target class vectors are converted from class indices corresponding to input vectors. This process generates a sparse matrix of vectors, with one 1 in each column, as indicated by indices. For example, if the $i$-th sample in training set is the $j$-th kind of plant, then we have one 1 on the $j$-th row of $i$-th column of  $\mathbf{M}$.

\begin{table*}
\caption[Training]{Details about the leaf numbers of different types of plants}
\begin{center}
\begin{tabular}[!htbp]{c | c | c | c }
\hline  Scientific Name(in Latin) & Common Name & training samples & number of incorrect recognition\\
\hline
\textit{Phyllostachys edulis} (Carr.) Houz.	& 	pubescent bamboo	 & 	58	 & 	0\\ 
\textit{Aesculus chinensis}			& 	Chinese horse chestnut	 & 	63	 & 	0\\ 
\textit{Berberis anhweiensis} Ahnendt		& 	Anhui Barberry	 	& 	58	 & 	0\\ 
\textit{Cercis chinensis}	 		& 	Chinese redbud	 	& 	72	 & 	1\\ 
\textit{Indigofera tinctoria} L.		& 	true indigo		& 	72	 & 	0\\ 
\textit{Acer Dalmatum}				& 	Japanese maple		& 	53	 & 	0\\ 
\textit{Phoebe zhennan} S. Lee \& F.N. Wei	& 	Nanmu	 		& 	60	 & 	1\\ 
\textit{Kalopanax septemlobus} (Thunb. ex A.Murr.) Koidz & castor aralia 	& 	51	 & 	0\\ 
\textit{Cinnamomum japonicum} Siebold ex Nakai	& 	Japan Cinnamon	 	& 	51	 & 	2\\ 
\textit{Koelreuteria paniculata} Laxm.		& 	goldenrain tree	 	& 	57	 & 	0\\ 
\textit{Ilex macrocarpa}			& 	holly	 		& 	50	 & 	0\\ 
\textit{Pittosporum tobira} (Thunb.) Ait. f.	& 	Japanese cheesewood	& 	61	 & 	1\\ 
\textit{Chimonanthus praecox} L.		& 	wintersweet	 	& 	51	 & 	2\\ 
\textit{Cinnamomum camphora} (L.) J. Presl	& 	camphortree	 	& 	61	 & 	3\\ 
\textit{Viburnum awabuki}	 		& 	Japanese Viburnum	& 	58	 & 	2\\ 
\textit{Osmanthus fragrans} Lour.	 	& 	sweet osmanthus	 	& 	55	 & 	5\\ 
\textit{Cedrus deodara} (Roxb.) G. Don		& 	deodar	 		& 	65	 & 	3\\ 
\textit{Ginkgo biloba} L.	 		& 	ginkgo, maidenhair tree	& 	57	 & 	0\\ 
\textit{Lagerstroemia indica} (L.) Pers.	& 	Crepe myrtle	 	& 	57	 & 	0\\ 
\textit{Nerium oleander} L.	 		& 	oleander	 	&	61	 & 	0\\ 
\textit{Podocarpus macrophyllus}  (Thunb.) Sweet& 	yew plum pine	 	& 	60	 & 	0\\ 
\textit{Prunus $\times$yedoensis} Matsumura	& 	Japanese Flowering Cherry & 	50	 & 	0\\ 
\textit{Ligustrum lucidum} Ait. f.	 	&  	Chinese Privet	 	& 	52	 & 	1\\ 
\textit{Tonna sinensis} M. Roem.		& 	Chinese Toon	 	& 	58	 & 	1\\ 
\textit{Prunus persica} (L.) Batsch	 	& 	peach	 		& 	50	 & 	2\\ 
\textit{Manglietia fordiana} Oliv.	 	& 	Ford Woodlotus	 	& 	50	 & 	3\\ 
\textit{Acer buergerianum} Miq. 		& 	trident maple	 	& 	50	 & 	1\\ 
\textit{Mahonia bealei}  (Fortune) Carr.		& 	Beale's barberry	& 	50	 & 	0\\ 
\textit{Magnolia grandiflora} L.	 	& 	southern magnolia	& 	50	 & 	0\\ 
\textit{Populus $\times$canadensis} Moench	& 	Carolina poplar	 	& 	58	 & 	3\\ 
\textit{Liriodendron chinense} (Hemsl.) Sarg.	& 	Chinese tulip tree	& 	50	 & 	0\\ 
\textit{Citrus reticulata} Blanco	 	& 	tangerine	 	& 	51	 & 	0\\
\hline
\end{tabular}
\label{training}
\end{center}
\end{table*}
\section{Experimental Result}
\label{result}

To each kind of plant, 10 pieces of leaves from testing sets are used to test the accuracy of our algorithm. Numbers incorrect recognition are listed in the last column of Table \ref{training}. The average accuracy is 90.312\%. 

Some species get a low accuracy in Table \ref{training}. Due to the simplicity of our algorithm framework, we can add more features to boost the accuracy.

We compared the accuracy of our algorithm with other general purpose (not only applicable to certain species) classification algorithms that only use leaf-shape information. According to Table \ref{accuracy}, the accuracy of our algorithm is very similar to other schemes. Considering our advantage respect to other automated/semi-automated general purpose schemes, easy-to-implement framework and fast speed of PNN, the performance is very good.
\begin{table}
\begin{center}
\caption[acc]{Accuracy comparison}
\begin{tabular}[!htb]{c | c }
\hline  Scheme & Accuracy\\
\hline proposed in \cite{A_Compuerized_Plant_Species_Recognition_System} & 71\% \\
 1-NN in \cite{Leaf_Recognition_Based_on_the_Combination_of_Wavelet_Transform_and_Gaussian_Interpolation} & 93\% \\
$k$-NN ($k=5$) in \cite{Leaf_Recognition_Based_on_the_Combination_of_Wavelet_Transform_and_Gaussian_Interpolation} & 86\% \\
RBPNN in \cite{Leaf_Recognition_Based_on_the_Combination_of_Wavelet_Transform_and_Gaussian_Interpolation} & 91\% \\
 MMC in \cite{Du2007} & 91\% \\
$k$-NN ($k=4$) in \cite{Du2007} & 92\%\\
MMC in \cite{Recognition_of_Leaf_Images_Based_on_Shape_Features_Using_a_Hypersphere_Classifier} & 92\% \\
BPNN in \cite{Recognition_of_Leaf_Images_Based_on_Shape_Features_Using_a_Hypersphere_Classifier} & 92\% \\
RBFNN in \cite{Du_ISNN} & 94\%\\
MLNN in \cite{Du_ISNN} & 94 \%\\
Our algorithm & 90\%\\
\hline
\end{tabular}
\label{accuracy}
\end{center}
\end{table}
 
The source code in MATLAB can be downloaded now from \url{http://flavia.sf.net}.

\section{Future Work}
\label{futurework}
Since the essential of the competitive function is to output the index of the maximum value in an array, we plan to let our algorithm output not only the index of maximum value, but also the indices of the second greatest value and the third greatest value. It is based on this consideration that the index of the second greatest value corresponds to the second top matched plant. So does the index of the third greatest value. Sometimes, maybe the correct plant is in the second or the third most possible plant. We are going to provide all these three possible answers to users. Further more, users can choose the correct one they think so that our algorithm can learn from it to improve its accuracy.


Other features are also under consideration. Daniel Drucker from Department of Psychology, University of Pennsylvania, suggested us to use Fourier Descriptors so that we can do some mathematical manipulations later. We are also trying to use other features having psychology proof that is useful for human to recognize things like the leaf, such as the surface qualities \cite{surface_qualities}.

Our plant database is under construction. The number of plants that can be classified will be increased.

\section{Conclusion}
\label{conclusion}
This paper introduces a neural network approach for plant leaf recognition. The computer can automatically classify 32 kinds of plants via the leaf images loaded from digital cameras or scanners. PNN is adopted for it has fast speed on training and simple structure. 12 features are extracted and processed by PCA to form the input vector of PNN. Experimental result indicates that our algorithm is workable with an accuracy greater than 90\% on 32 kinds of plants. Compared with other methods, this algorithm is fast in execution, efficient in recognition and easy in implementation. Future work is under consideration to improve it.

\section*{Acknowledgements}
Prof. Xin-Jun Tian, Department of Botany, School of Life Sciences, Nanjing University provided the lab and some advises for this research. 

Yue Zhu, a master student of Department of Botany, School of Life Sciences, Nanjing University, helped us sampling plant leaves.

Ang Li and Bing Chen from Institute of Botany, Chinese Academy of Science, provided us some advises on plant taxonomy and searched the scientific name for plants.

Shi Chen, a PhD student from School of Agriculture, Pennsylvania State University, initiated another project which inspired us this research.

The authors also wish to thank secretary Crystal Hwan-Ming Chan, for her assistance to our project.

\bibliographystyle{IEEEtran}
\bibliography{flavia}

\end{document}